\documentclass[letterpaper, 10 pt, journal, twoside]{IEEEtran}  
\usepackage[T1]{fontenc}
\usepackage{pgfplots}
\usepackage{subcaption}
\usepackage{graphicx}
\usepackage{multirow}
\usepackage{pgfplotstable}
\usepackage{tabularx}
\usepackage{algorithm}
\usepackage{algpseudocode}
\usepackage{amsmath}
\usepackage{cite}
\usepackage{hyperref}
\usepackage{enumitem}
\usepackage{tikz}
\usetikzlibrary{shapes, arrows, positioning, fit}
\usepackage{graphicx}
\pgfplotsset{compat=1.18} 
\usetikzlibrary{pgfplots.groupplots} 

\IEEEoverridecommandlockouts                              

\title{\LARGE \bf
OVITA: Open-Vocabulary Interpretable Trajectory Adaptations
}

\author{Anurag Maurya$^{1}$, Tashmoy Ghosh$^{1}$, Anh Nguyen$^{2}$ and Ravi Prakash$^{1}$ 

\thanks{Manuscript received: March, 21, 2025; Revised June, 28, 2025; Accepted August, 13, 2025.}
\thanks{This paper was recommended for publication by Editor Olivier Stasse upon evaluation of the Associate Editor and Reviewers' comments.
} 
\thanks{$^{1}$First Author, Second Author, and Fourth Author are with Robert Bosch Centre for Cyber-Physical Systems, Indian Institute of Science, Bangalore, India
        {\tt\footnotesize (anuragm1,tashmoyg,ravipr)@iisc.ac.in}}%
\thanks{$^{2} $Third Author is with the Department of Computer Science, University of Liverpool, England
        {\tt\footnotesize anh.nguyen@liverpool.ac.uk}}%
\thanks{Digital Object Identifier (DOI): see top of this page.}
}

\begin{document}
\maketitle


\begin{abstract}


Adapting trajectories to dynamic situations and user preferences is crucial for robot operation in unstructured environments with non-expert users. Natural language enables users to express these adjustments in an interactive manner. We introduce OVITA, an interpretable, open-vocabulary, language-driven framework designed for adapting robot trajectories in dynamic and novel situations based on human instructions. OVITA leverages multiple pre-trained Large Language Models (LLMs) to integrate user commands into trajectories generated by motion planners or those learned through demonstrations. OVITA employs code as an adaptation policy generated by an LLM, enabling users to adjust individual waypoints, thus providing flexible control. Another LLM, which acts as a code explainer, removes the need for expert users, enabling intuitive interactions. 
The efficacy and significance of the proposed OVITA framework is demonstrated through extensive simulations and real-world environments with diverse tasks involving spatiotemporal variations on heterogeneous robotic platforms such as a KUKA IIWA robot manipulator, Clearpath Jackal ground robot, and CrazyFlie drone.
\end{abstract}


\IEEEpeerreviewmaketitle

\section{INTRODUCTION}


Robotic systems have increasingly permeated diverse domains, from industrial automation to service robotics, demanding efficient trajectory generation and adaptation techniques. A fundamental challenge in this context lies in enabling robots to generalize in dynamic and unstructured environments. Large Language Models (LLMs) have gained attention for their reasoning abilities, driving their use in robotics, particularly for high-level task planning in embodied AI and human-robot collaboration \cite{wang2024survey,song2023llm,singh2023progprompt}.  However, LLMs are underutilized in low-level control, often relying on pretrained skills and motion primitives. Methods like VoxPoser \cite{huang2023voxposer} generate affordances, cost functions, and 3D cost maps, while Language to Rewards \cite{yu2023language} designs reward functions. Recent works \cite{kwon2024language} has shown LLMs can generate zero-shot dense trajectories, but their outputs remain inferior to human demonstrations and are limited to simple shapes and linear interpolations.

Traditionally, robot trajectories are either planned from scratch using sampling-based methods (e.g. Rapidly-Exploring Random Trees (RRT), RRT* \cite{C14} and MPC \cite{rawlings2017model}) or adapted from previously learned demonstrations. While the former lacks online reactivity to uncertain situations, the latter learns adaptive and reactive motion plans through Learning from Demonstrations (LfD)  \cite{C15} that are inherently robust to uncertainties and changes in the environment. Robot learning from demonstrations allows humans, regardless of expertise, to transfer task knowledge to robots \cite{C17}, eliminating the need for tedious, error-prone manual programming. However, a key challenge remains in generalizing learned behaviors to new situations. Moreover, a high number of samples is typically needed for each task.

\begin{figure}
    \centering
    \includegraphics[width=\linewidth]{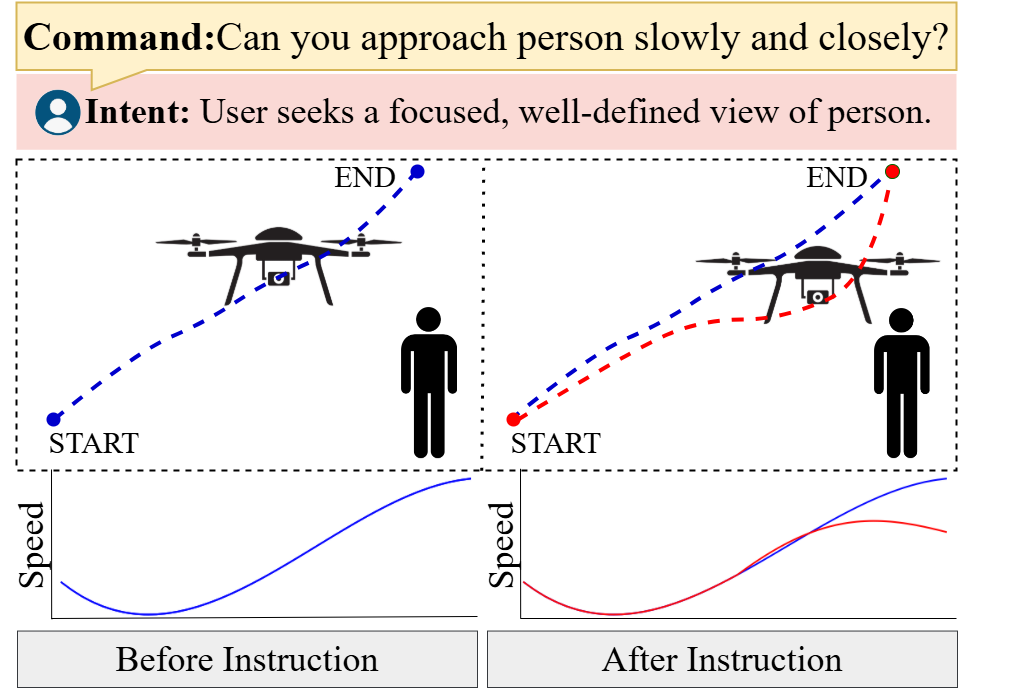}
    \caption{Traditional planning algorithms often overlook user intent. OVITA enables intuitive adjustments through language commands. The blue trajectory represents the initial path, while the red one reflects sample OVITA output.}
    \label{fig:method}
    \vspace{-15pt}
\end{figure}


 We propose OVITA, a user-friendly robot trajectory adaptation method that leverages LLMs for interactive and interpretable adjustments via natural language. By eliminating complex task parameterization, OVITA unifies trajectory adaptations under a single framework. Natural language instructions enable users to convey intent intuitively, allowing robots to dynamically adjust to task variations.

For example, a planner may generate a trajectory for a drone to surveil a region. However, if the user intends to have a focused and well-defined view of an individual, they can issue a command like, "Can you approach person slowly and closely?" (Fig. \ref{fig:method}). In this context, natural language serves as a medium for the user to express specific preferences. Similarly, in object transportation tasks, a predefined trajectory might guide the robot in loading items onto a table. However, if the user needs more time to add additional items, they could issue a command like, "Stop near the table for 10 more timesteps," allowing the robot to adapt dynamically to the revised requirements. This flexibility highlights the value of combining human instructions with trajectory adaptation.

Natural language provides an intuitive medium for users to express their intent, but its open-vocabulary nature introduces unique challenges. Unstructured commands can result in inconsistent interpretations. For example, the instruction "Stay at a larger distance from the cup" may be understood differently by users—one might interpret it as a slight adjustment to only nearby points, while another might see it as requiring a shift to all the waypoints. By offering corrections when adaptations deviate from expectations, users can iteratively refine the robot's actions, leading to more personalized and contextually appropriate adjustments. Transparency and interpretability are equally vital in this adaptation process \cite{anjomshoae2019explainable}. Black-box systems that obscure trajectory modifications make it difficult for users to assess whether their instructions have been correctly implemented. On the other hand, clear visualization of adjustments empowers users to evaluate the outcomes, identify discrepancies, and refine their commands.

Starting with a trajectory from a planner or human demonstration, our method adjusts low-level actions using natural language commands. OVITA eliminates the reliance on complex mathematical formulations and prioritizes intuitive, user-driven adjustments. Our method adjusts key waypoints while maintaining trajectory smoothness. It interprets natural language as specific modifications and encodes them into actionable updates using code as an adaptation policy.

\noindent OVITA offers the following key contributions:
\begin{itemize}
    \item It uses pre-trained LLMs to interpret open-vocabulary instructions for spatiotemporal trajectory adaptations in a multi-agent framework.
    \item It supports exact numerical changes (e.g., "Go left by 0.2"), open-ended commands (e.g., "Move in a spiral"), and multi-step adjustments with feedback.
    \item {The integration of a formal, QP-based optimization module that translates LLM-generated adaptations into smooth, safe, and physically-feasible trajectories.}
    \item Real-world experimental validation of framework on heterogeneous robotics platforms and tasks.
\end{itemize}
To the best of our knowledge, OVITA is the first interpretable framework for trajectory adaptation with open-vocabulary instruction support. The paper is structured as follows: Section II reviews related literature, Section III outlines the proposed methodology, Section IV presents results and discussions, and Section V covers limitations and conclusions.

\section{RELATED WORKS}

\subsection{Trajectory adaptation in Robotics}
In new or changing scenarios, robot motion plans must adapt to sensory feedback—a key research area in Imitation Learning \cite{osa2018algorithmic}. By enhancing the policy with task parameterization \cite{C16}, robots can generalize their learned knowledge to different variations of the same task, allowing robots to adapt to new scenarios. This can be popularly achieved by Dynamic Movement Primitives (DMPs) \cite{DMP}, which uses second-order stable attractor dynamics with adjustable spatial targets and temporal execution. Alternatively, skill generalization has been formulated with ProMPs \cite{ProMP} and KMPs \cite{KMP}. While ProMP uses basis functions and Gaussian conditioning to adapt trajectories to any waypoints while maintaining temporal and DoF correlations, KMP adapts these trajectories using kernel-based temporal correlations. By encoding probability distributions of trajectories concerning task-relevant reference frames, TP-GMM(Task-Parameterized Gaussian Mixture Model) enables motion generation through TP-GMM + GMR/GPR \cite{C18} with improved out-of-distribution targets. Despite their utility, these methods require task knowledge, parameter tuning, and extensive expert demonstrations, limiting scalability for non-experts.
OVITA seamlessly adapts to out-of-distribution targets, modifying waypoints solely through natural language instructions. In addition, we can add a new task and generate extra waypoints originally not present in the original trajectories. 

\subsection{Large Language Models for Robotics}
Recent advancements in LLMs have facilitated their integration into robotic task planning. The key works in high-level task planning include the following. PaLM-E \cite{driess2023palm} introduces embodied language models that connect language with real-world sensor data, using multimodal inputs and outputs high-level plan. LLaRP \cite{szot2023large} uses an LLM-based reinforcement learning policy to translate text and visual inputs into a plan of predefined actions. RobotGPT \cite{jin2024robotgpt} uses ChatGPT to iteratively refine code to generate data for training agents. SayCan \cite{C8} combines LLM-based task grounding with value functions for action feasibility, while Code-as-Policies \cite{C9} pioneered generating interpretable, executable code with predefined motion primitives. Unlike end-to-end models, this approach offers debuggable logic, linking language, and robotic control.  To incorporate user preferences into high-level robotic plans, \cite{huanginner} uses a closed-loop strategy with feedback like success detection, scene description, and human interaction. LLMs have also been used to generate low-level trajectories, translating task descriptions into movement sequences \cite{kwon2024language}. Their approach is limited to simple trajectories and tabletop tasks, while ours adapts complex trajectories to user preferences and supports diverse robot dynamics and morphologies.

\subsection{Trajectory Adaptation with Natural Language}
Natural language is a powerful interface for robots, but linking it to low-level motion plans is challenging. Traditionally, representing human-robot interactions through language has required either rigid instructions \cite{C1} or complex mathematical methods to model probabilities of actions and targets \cite{C2}. LILAC \cite{C12} enables online trajectory corrections through natural language inputs in shared autonomy. Recent advances in NLP foundation models like BERT \cite{C19} have enabled end-to-end learning with deep embeddings. Prior work has leveraged LLMs for either optimizing cost maps via motion planners \cite{C4} or mapping input to output trajectories based on geometric object poses and LLM-based language embeddings \cite{C5, C6}. Notably, the failure analysis of LaTTe \cite{C6} by ExTraCT\cite{C7} highlighted that embedding models often generate similar embeddings for sentences with opposing meanings but with high lexical similarity. Additionally, prior approaches rely on templated, qualitative instructions and lack support for precise corrections (e.g., "Go left by 0.3") or multi-instruction commands. These limitations motivate our development of a robust, language-driven trajectory adaptation pipeline.
OVITA incorporates precise numerical adjustments and multi-instruction commands. Additionally, It handles open-ended directives, such as: “Set the new goal position as the midpoint between the box and sofa, and modify the trajectory smoothly to reach it,” or “Move slightly closer to the punching bag and follow a trapezoidal velocity profile.”. This capability sets our method apart.


{The most similar works to ours are LaTTe (Language Trajectory Transformers) \cite{C6} and ExTraCT \cite{C7}. LaTTe combines language embeddings and spatial representations in a multi-modal, end-to-end framework. However, it requires extensive training data, is non-explainable, and lacks support for complex and open-vocabulary interactions. ExTraCT uses language embeddings to map deformation functions but depends on handcrafted features, limiting its generalizability.} In contrast, OVITA is a training-free(zero-shot) method for complex, multi-step numerical adaptations. It requires no fine-tuning, is fully interpretable, and supports open-vocabulary, monologue-based interactions. Additionally, feedback support facilitates iterative refinement.

\section{METHODOLOGY}

\begin{figure}
    \centering
    \includegraphics[width=\linewidth]{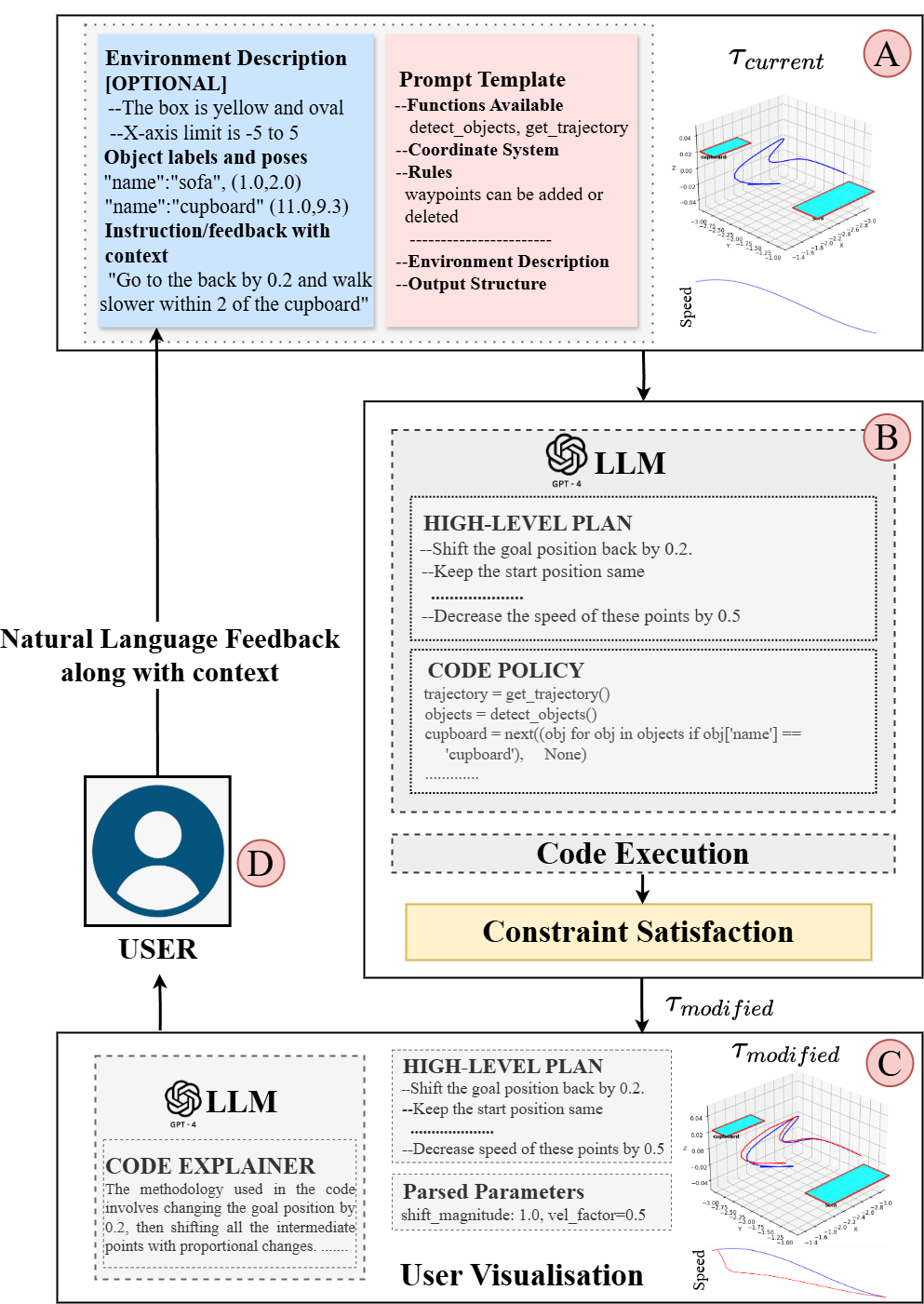}
    \caption{\textbf{OVITA Block Diagram:} A) Prompt grounding from instruction and environment description.
B) LLM generates HLP and code, followed by execution and constraint satisfaction. C) 
Visualization with code explanation.
D) User feedback for refinement.}
    \label{fig:block_diagram}
    \vspace{-12.5pt}
\end{figure}

\subsection{{Language-based Adaptation Pipeline}}
Given an initial trajectory  
$\tau = \{(x_i, y_i, z_i, v_i)\}_{i=1}^N$,  
consisting of $N$ waypoints, where $(x_i, y_i, z_i)$ represent the coordinates and $v_i$ the corresponding speed.  
We also consider a user instruction $L_{\text{instruct}}$ and environment observations  
$\{\text{Position}(O_i), \text{Dimension}(O_i), \text{Label}(O_i)\}_{i=1}^m$ for $m$ objects.  
These observations may optionally include an environmental description $E_d$.  
Our goal is to construct a mapping  
$g: \tau_{\text{initial}} \rightarrow \tau_{\text{modified}}$  
that adapts the trajectory according to $L_{\text{instruct}}$.  
We leverage LLMs to achieve this adaptation. The adaptation process is structured in four phases. 
prompt grounding, code-policy execution, visualization, and feedback integration. 

\subsection*{First Phase: Prompt Grounding} 
{
\noindent This phase translates user instructions into a grounded prompt (Fig.~\ref{fig:block_diagram} A) using a task-agnostic template that provides context for LLM decisions and structured code generation. The template includes core components common to all adaptations. \textit{1) Environment configuration:} A defined coordinate system and a placeholder for environment descriptors to provide both spatial and semantic context. \textit{2) Trajectory Characteristics:} Rules for smooth transitions and to allow for adding or removing waypoints, etc. \textit{3) Function Definitions:} The prompt includes functions like detect\_objects() to retrieve object positions and get\_trajectory() for trajectory data.\textit{ 4) Examples:} Two examples demonstrate the response format for a high-level plan, not the code policy. These examples are kept the same across all tasks. \\ \textbf{Prompt Grounding:} The user observes the initial trajectory and issues a natural language command.We add environmental context using semantic descriptions, either generated by a vision–language model (VLM) or provided by the user. The VLM describes objects in the scene, including their names, properties (e.g., color), and layout. An object detector then extracts their 3D positions and sizes. (Details in \ref{real-world-experiments}). The user's instruction, scene descriptions, and object positions are then structured within the prompt template, resulting in a physically grounded prompt.  prompts are provided in the prompts.py file of the codebase.
}

\subsection*{Second Phase: Code-Policy generation and Execution}
\begin{figure}
    \centering
    \includegraphics[width=0.85\linewidth]{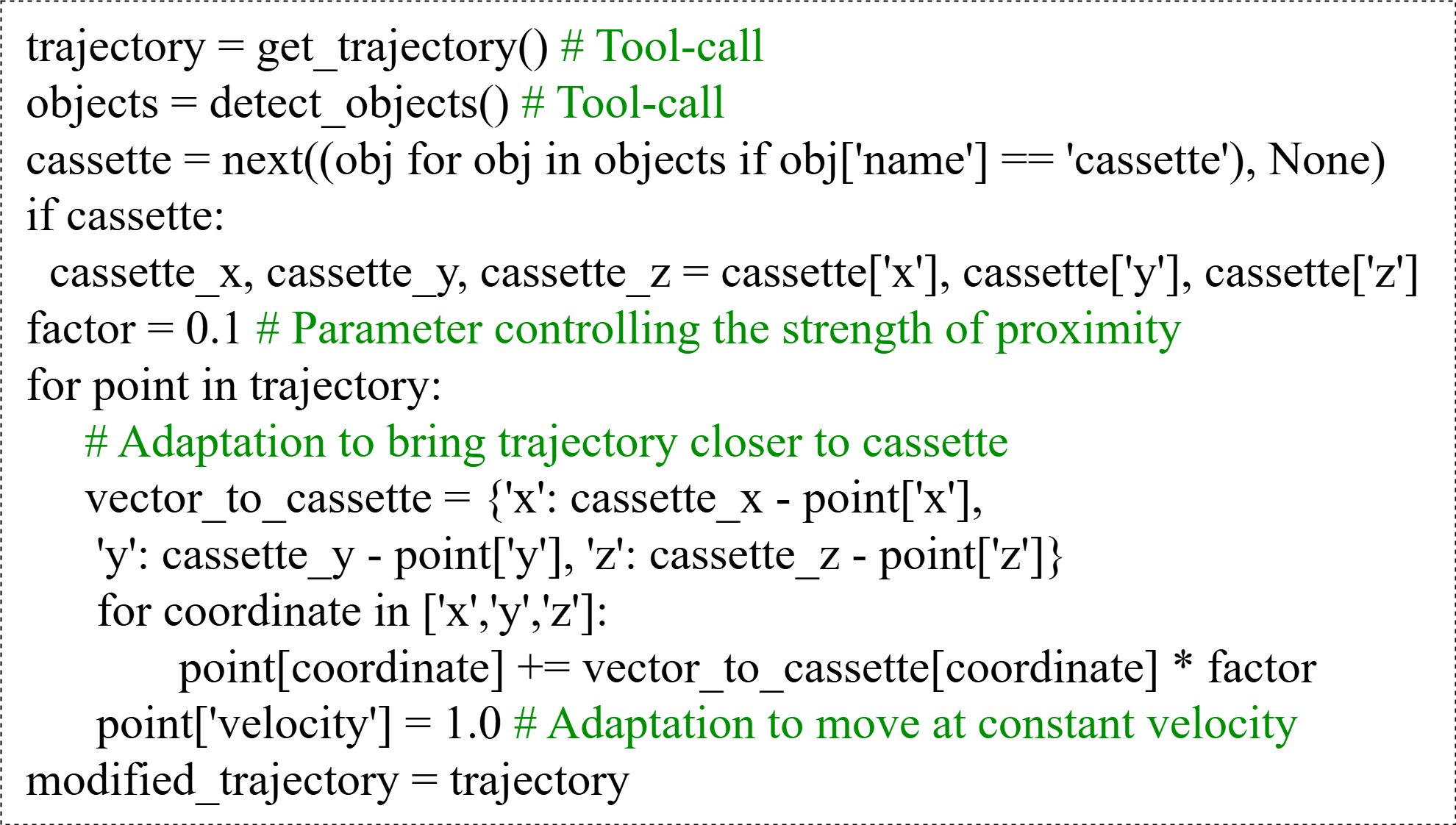}
    \caption{Code as an adaptation policy: the instruction is "Move closer to the cassette while maintaining a constant speed throughout."}
    \label{fig:code_policy}
    \vspace{-15pt}
\end{figure}

\noindent {In this phase, the LLM processes a unified prompt to generate a dictionary with both a high-level plan and Python code in a single generation step }(Fig.~\ref{fig:block_diagram} B). Empirically, we observed that this method leads to higher consistency compared to separate generations. The high-level plan specifies the precise steps required to align the trajectory with user instructions. This plan guides the LLM in generating Python code, which transforms input trajectory data (position and velocity sequences) into adapted output sequences. The code bridges high-level directives and low-level execution, acting as an adaptation policy(Fig.\ref{fig:code_policy}). AST(Abstract Syntax Trees) parsing is employed to refine the generated code, extract key parameters, and remove unnecessary elements. The resulting code is executed in a Python interpreter, producing the adapted trajectory. An optional constraint satisfaction module (CSM) validates the trajectory, ensuring the environment and robot-specific constraints (Details in \ref{subsection:CSM}).

\subsection*{Third Phase: User Visualization}
\noindent In this phase, the user is shown a comparative visualization (trajectory plots) of the adapted and initial trajectories, along with access to the high-level plan, and parsed variables (Fig.\ref{fig:block_diagram} C). This enables users to provide well-informed feedback. To mitigate potential interpretation challenges for non-expert users, a code explanation module was integrated. This module utilizes an LLM to generate accessible explanations from the code, high-level plan, and extracted parameters. { \textbf{Code-explanation prompt} explains code-adaptation policy by: 1) summarizing the methodology used; 2) detailing the function and values of key hyperparameters; and 3) outlining logical assumptions underlying trajectory adjustments.}

\subsection*{Fourth Phase: Feedback integration}

\begin{figure}
    \centering
    \includegraphics[width=\linewidth]{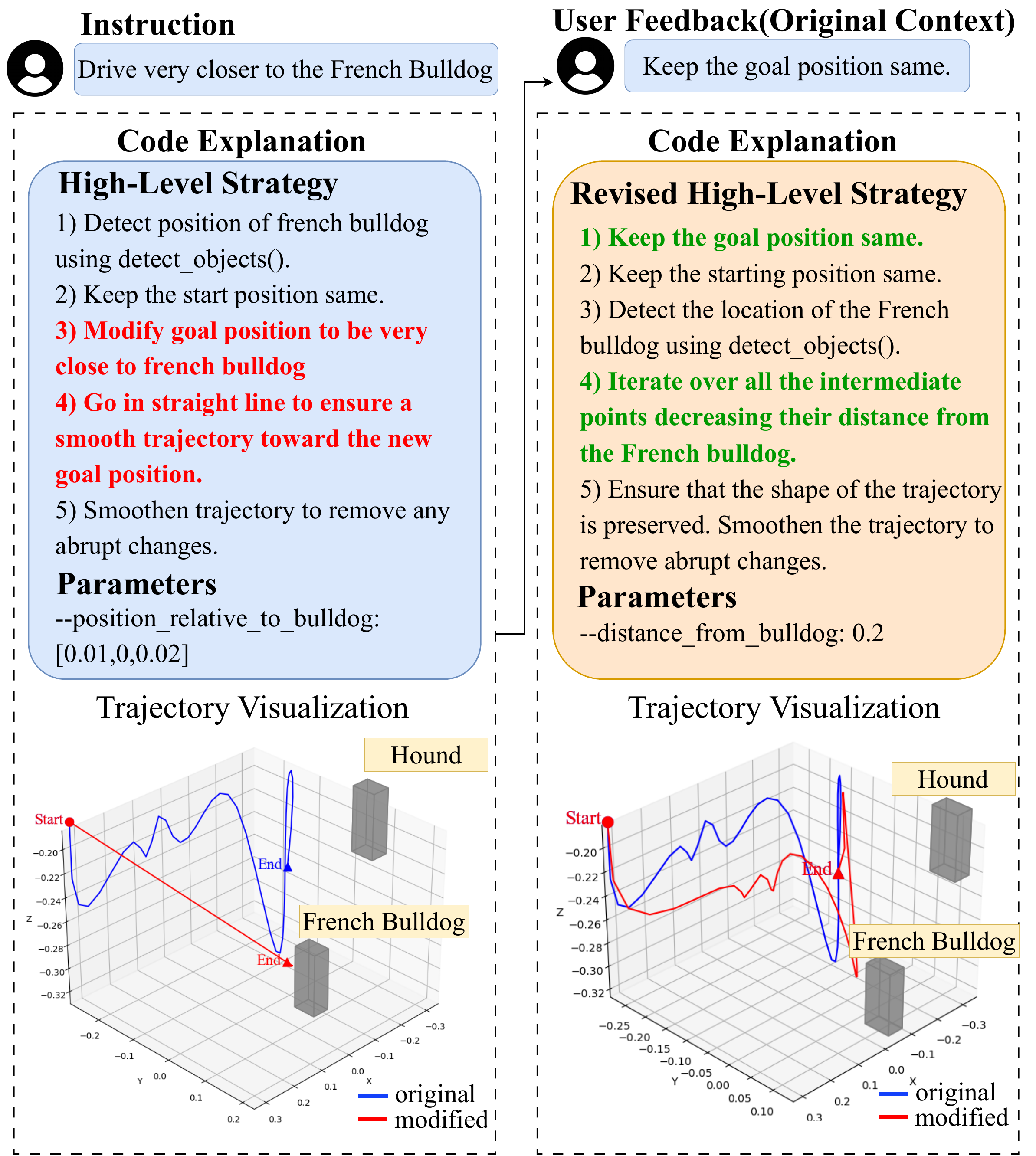}
    \caption{ Visual illustration of our approach to mitigate LLM hallucinations: Interactive trajectory visualization helps identify errors (red) and enables targeted user feedback for refinement. Code explanations with parameters aid in identifying missteps to produce updated strategies (green)}
    \label{fig:feedback}
    \vspace{-12.5pt}
\end{figure}

\noindent This phase introduces a crucial feedback loop(Fig.\ref{fig:block_diagram} D). Natural language ambiguity and user diversity can cause LLM hallucinations. To address this, natural language feedback is incorporated, enabling iterative corrections. As illustrated in Fig. \ref{fig:feedback}, this feedback loop effectively refines the high-level plan, mitigating LLM hallucinations and enhancing adherence to user preferences. We observed ambiguity in user feedback, as it could target either the initial or current trajectory. To avoid confusion, we implemented a feedback context selector, offering "original" or "current" options. Choosing "Original" appends new feedback to the original instruction, applying the resulting code policy to the original trajectory. Selecting "current" applies the code policy to the previously modified trajectory.

\subsection{{Constraint Satisfaction Module}}
{
\label{subsection:CSM}

\begin{figure*} [h]
        \centering
    \includegraphics[width=\linewidth]{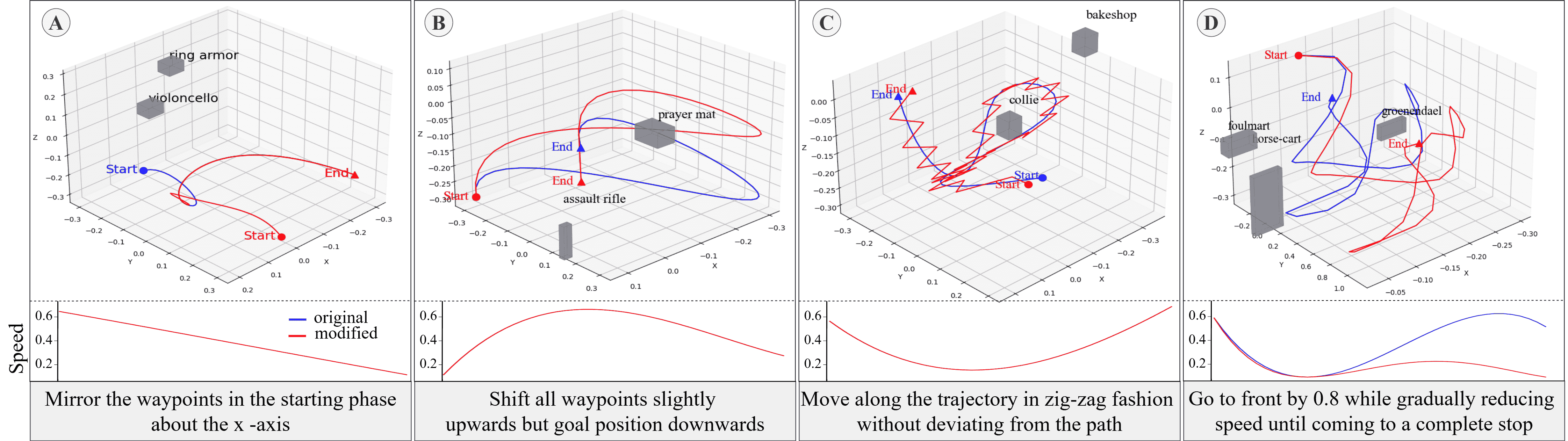}
    \caption{\textbf{Results on open-vocabulary instructions:} The original trajectory is represented in blue, while the modified trajectory using our approach is shown in red. }
    \label{fig:anecdotal_res}
    \vspace{-0.2in}
\end{figure*}

We formulate the trajectory optimization as a convex Quadratic Program (QP) over a sequence of \( T \) 3D waypoints and their speed \( \mathbf{x} = [\mathbf{x}_0^\top, \mathbf{x}_1^\top, \dots, \mathbf{x}_{T-1}^\top]^\top \in \mathbf{R}^{4T} \).  The objective penalizes deviation from the reference trajectory \( \mathbf{x}^{\mathrm{ref}} \) and encourages smoothness via first-order differences.

\begin{equation}
\min_{\mathbf{x} \in \mathbf{R}^{4T}} \quad \lambda_{\text{dev}} \|\mathbf{x} - \mathbf{x}^{\mathrm{ref}}\|_2^2 + \lambda_{\text{smooth}} \|D_1 \mathbf{x}\|_2^2
\end{equation}

where \( \lambda_{\text{dev}} > 0 \) and \( \lambda_{\text{smooth}} > 0 \) are weights, and \( D_1 \) is the block-diagonal first-order difference matrix.

\noindent \textbf{Constraints.} We impose the following linear inequality constraints, where \( G \) and \( \mathbf{h} \) encode workspace and obstacle avoidance terms mentioned in Eq.~\ref{eq:cuboidal_ws},Eq.~\ref{eq:spherical_ws} , Eq.~\ref{eq:obs_avoid} and Eq.~\ref{eq:velocity}:
\begin{equation}
G\mathbf{x} \leq \mathbf{h}
\label{eq:inequality}
\end{equation}

\begin{enumerate}[leftmargin=*]
    \item \textbf{Workspace constraints:} $\delta$ is safety margin parameter.
    \begin{itemize}
        \item \emph{Cuboidal bounds (for drones and ground robots):}
        \begin{equation}
        \label{eq:cuboidal_ws}
        x_i^{\min} + \delta \leq \mathbf{x}_t^{(i)} \leq x_i^{\max} - \delta, \quad i \in \{x, y, z\}
        \end{equation}

        \item \emph{Spherical bounds (for manipulators):}
        \begin{equation}
        \label{eq:spherical_ws}
        R_{\min}^2 \leq \|\mathbf{x}_t - \mathbf{c}\|_2^2 \leq R_{\max}^2
        \end{equation}
    \end{itemize}

    \item \textbf{Obstacle avoidance:} Each obstacle is approximated by a bounding sphere of radius \( R_{\text{obs}} \), centered at \( \mathbf{o} \). 
    \begin{equation}
    \label{eq:obs_avoid}
    \|\mathbf{x}_t - \mathbf{o}\|_2^2 \geq R_{\text{obs}}^2 + \delta^2
    \end{equation}

Spherical workspace and obstacle avoidance constraints are linearized via first-order Taylor expansion around the reference trajectory \( \mathbf{x}^{\mathrm{ref}} \). In our pipeline, the reference is the LLM-modified trajectory reflecting user intent. The CSM then refines this path locally for safety, keeping the final trajectory near $\mathbf{x}^{\mathrm{ref}}$. This makes linear constraint approximations both accurate and well-suited.

\item \textbf{Velocity constraint:}
\begin{equation}
\label{eq:velocity}
     v_i \leq v_{\text{max}} 
\end{equation}

\item \textbf{(Optional) Equality Constraints.} To enforce fixed start and/or goal configurations:
\begin{equation}
\label{eq:equality_constraints}
A\mathbf{x} = \mathbf{b} \quad \text{:} \quad \mathbf{x}_0 = \mathbf{x}^{\mathrm{ref}}_0, \quad \mathbf{x}_{T-1} = \mathbf{x}^{\mathrm{ref}}_{T-1}
\end{equation}
\end{enumerate}

\noindent\textbf{QP Formulation} We formulate P and Q matrices as
\begin{equation}
\label{eq:P_Q}
P = 2(\lambda_{\text{dev}} I + \lambda_{\text{smooth}} D_1^\top D_1), \quad \mathbf{q} = -2\lambda_{\text{dev}} \mathbf{x}^{\mathrm{ref}}
\end{equation}
From Eq~\ref{eq:inequality}, Eq~\ref{eq:equality_constraints},and Eq~\ref{eq:P_Q}, the full QP becomes:
\[
\begin{aligned}
\min_{\mathbf{x} \in \mathbf{R}^{4T}} & \quad \frac{1}{2} \mathbf{x}^\top P \mathbf{x} + \mathbf{q}^\top \mathbf{x} \\
\text{s.t.} & \quad G \mathbf{x} \leq \mathbf{h},
            \quad A \mathbf{x} = \mathbf{b} \quad (\text{optional}).
\end{aligned}
\]
This problem is solved using a standard convex QP solver. $\lambda_{dev}$ and $\lambda_{smooth}$ are chosen through Optuna~\cite{optuna_2019} by minimizing the deviation cost. The search employs the Tree-structured Parzen Estimator (TPE) over a logarithmic range \( 10^{-3} \) to \( 10^{1} \) with 50 trials. 
Our framework’s modularity allows seamless integration of custom constraint modules.
}

\section{EXPERIMENTS AND RESULTS}

This section presents a detailed evaluation of our method. Our experiments address the following key points: \\
     \textbf{Q1}: How can our method handle free-form natural language commands (open-ended, not restricted by predefined syntax or rigid formats) for spatiotemporal manipulation of the waypoints? \\
\textbf{Q2}: To what extent is user feedback essential in achieving the desired adaptations? Additionally, how does the interpretability of the proposed approach enable an effective understanding of adaptations?
   
\subsection{Dataset}

Each data sample includes an initial trajectory (\(\tau_0\)), environmental observations (Object \(\text{Label}(O_i)\), Positions \( \text{P}(O_i)\), Dimensions \( \text{D}(O_i)\)), an optional linguistic description (\(E_d\)), and a natural language instruction (\(L_{\text{instruct}}\)). Unlike previous methods {\cite{C5,C6,C7}} that rely on point objects, we represent objects as cuboids for more realistic constraint satisfaction. We use a subset (75 trajectories) of the LaTTe (Language Trajectory Transformer) \cite{C6} dataset. The original LaTTe dataset focuses on three instruction types: (1) Cartesian waypoint changes (e.g., "Go to the left"), (2) speed adjustments (e.g., "Go slower"), and (3) positional shifts relative to objects (e.g., "Drive closer to the sofa"). The dataset has template-based changes over precise numbers and uses single-instruction templates with a fixed vocabulary. To overcome these limitations, we create a subset with open-vocabulary instructions. Specifically, we include complex commands with numerical specifications. 
(e.g., ``Go left by 20''), combined instructions 
(e.g., ``Go to the front by 0.8 while gradually reducing speed until coming to a complete stop''), 
and long open-ended commands (e.g., ``Move along the trajectory in zig-zag fashion without deviating from the path'').
Fig.~\ref{fig:anecdotal_res} shows examples from the extended subset. 

To evaluate our approach to robot deployment under constraints, we curated trajectories from simulated and real-world environments. They include wheeled, aerial, and manipulator robots.
The complete dataset includes 230 trajectories: LaTTe subset (75), extended subset (75), and robot subset (80), evaluated across three LLMs { (gpt-4o-2024-11-20 \cite{hurst2024gpt}, claude-3-opus-20240229 \cite{anthropic2024claude}, and gemini-1.5-pro-002 \cite{team2023gemini}) with varied feedback. }

\subsection{Evaluation Design}

Due to the open-vocabulary nature of the instructions and the multiple possible correct adjustments for any given trajectory, designing a comprehensive evaluation metric for all instructions is highly challenging. Metrics like RMSE require a supervised trajectory as a reference, which isn't feasible for every open-vocabulary instruction. Moreover, as noted by \cite{C7}, these metrics don’t fully capture the nuances of trajectory adaptations. For instance, two trajectories—one approaching and the other receding from a reference by the same distance—can yield identical RMSE values, yet very different semantic meanings. We resort to a user study to evaluate the results. 

We conducted 690 evaluations with 30 participants (15 robotics experts and 15 non-experts). The evaluation consists of two stages. In the first stage, participants receive a trajectory adapted as per dataset instructions. They can refine it with feedback and save the results. In the second stage, a different participant assessed the adapted trajectories, along with feedback and code explanations, to determine their semantic correctness. This eliminates self-confirmation bias, ensuring an unbiased evaluation. Trajectories were presented in an interactive 3D environment, allowing evaluators to adjust the view. They were assessed on a 5-point Likert scale, from (1) Completely wrong to (5) Completely correct. Participants also rated the usefulness of code interpretations for understanding trajectory adaptation on the same scale(interpretability score). This scale serves as a performance metric for our method, with task success defined as scores of 4 or 5, while scores of 1, 2, or 3 are considered failures.

\subsection{Results and Discussion}

\begin{figure}[t]
    \centering
    \includegraphics[width=0.9\linewidth]{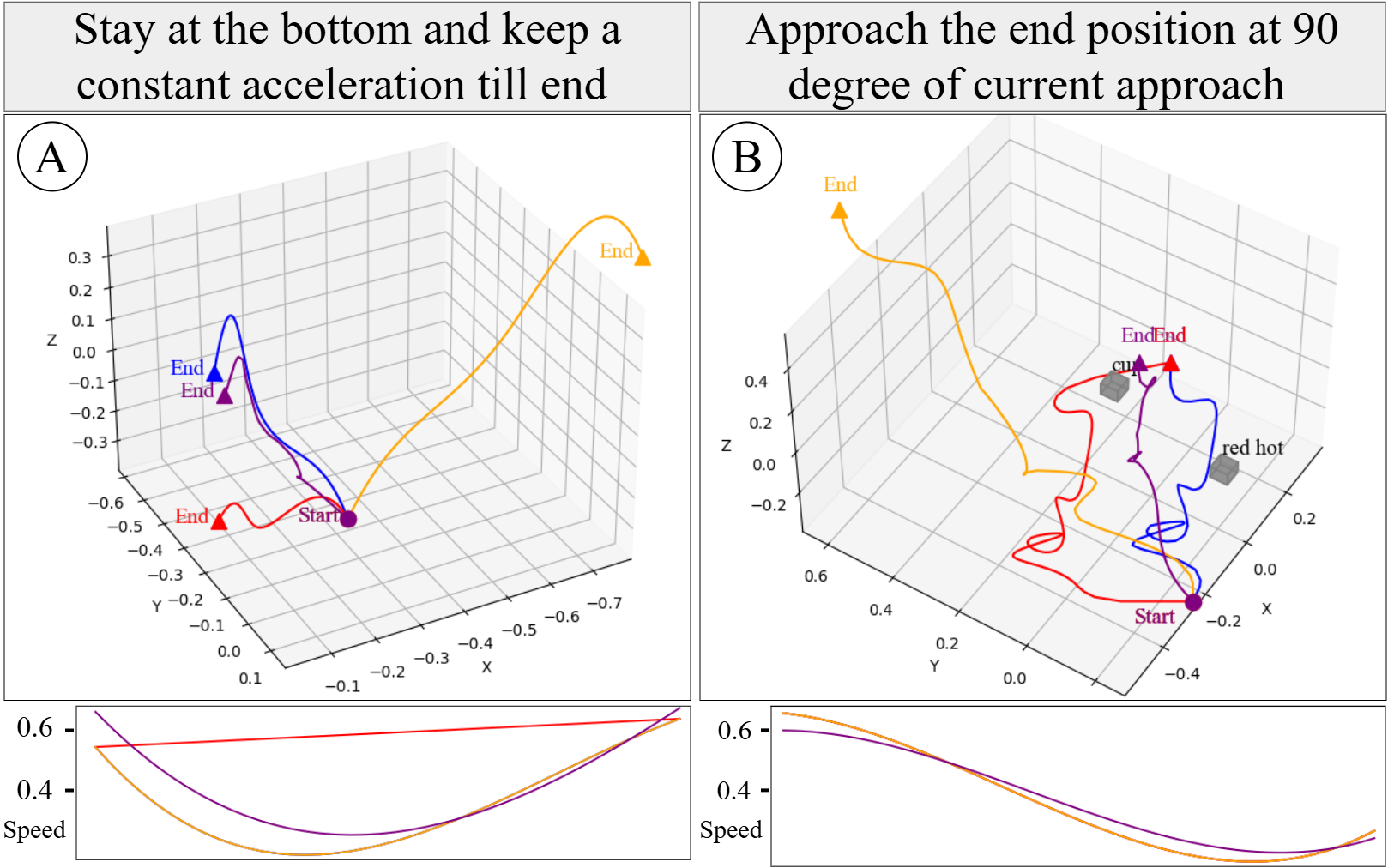}
    \caption{{OVITA, LaTTe, and ExTracT Trajectory Comparison on extended subset: The initial trajectory is depicted in blue. OVITA, LaTTe, and ExTraCT trajectories are shown in Red, Purple, and Yellow, respectively. Due to overlap, speed plots may not be distinctly visible.}}         
    \label{fig:latte_compare_qualitative}
    \vspace{-10pt}
\end{figure}

\begin{figure}[t]
    \centering
    \includegraphics[width=0.8\linewidth]{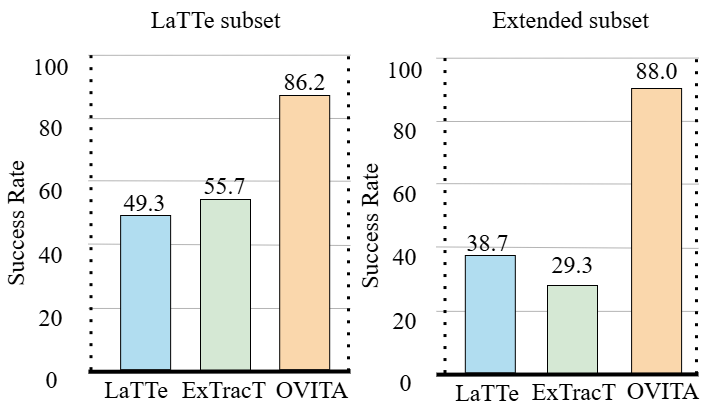}
    \caption{ {Performance Comparison between OVITA, LaTTe and ExTraCT on LaTTe subset and extended subset}}
    \label{fig:latte_compare}
    \vspace{-0.25in}
\end{figure}

\begin{figure*}[h]
        \centering
    \includegraphics[width=\linewidth]{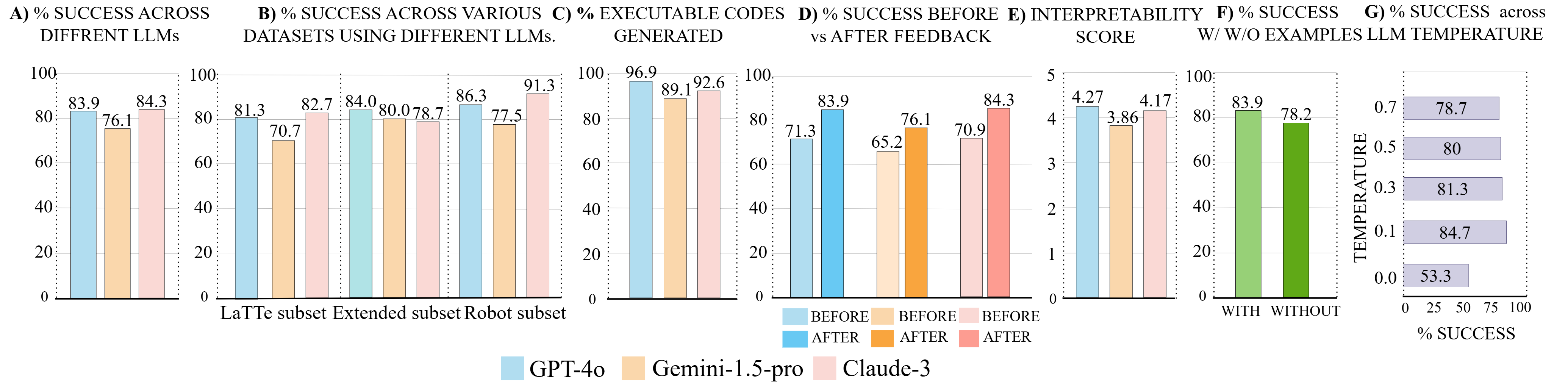}
    \caption{
{Evaluation of OVITA: (A) across various LLMs, (B) dataset subsets, (C) executable code generation rates, (D) feedback impact on success, (E) interpretability scores, (F) ablation w/ and w/o examples, and (G) LLM temperature variations}}
    \label{fig:quant}
    \vspace{-15pt}
\end{figure*}

 Fig.\ref{fig:quant} illustrates the success rates of trajectories across various dataset subsets, comparing outcomes with and without feedback while Fig. \ref{fig:anecdotal_res} presents the qualitative findings. 

\noindent\textbf{Evaluating Q1}: Our approach achieved an 81.4\% average success rate, excelling on both the extended and Robot subsets (Fig.~\ref{fig:quant} B), which involve open-vocabulary, freeform, and complex instructions. Notably, the high success rate on the Robot subset confirms our method’s ability to generate constraint-compliant trajectories suited for real-world scenarios. A deeper analysis revealed the following insights: \\
\textbf{1) }Unlike prior methods focused on transforming existing waypoints, our approach can handle both generation and transformation. For instance, the command “create a spiral in the middle of the trajectory” requires generating new waypoints to form a spiral while maintaining alignment with the original trajectory(Fig.\ref{fig:sim_res}). Thus retaining the first half of the trajectory, generating the spiral, and then continuing the original trajectory. \textbf{2)} Our code-based approach treats waypoints as independent, allowing flexible trajectory modifications. This enables diverse user intents. For example, the instruction “Shift all waypoints slightly upwards but goal position downwards” allows local adjustments to the goal waypoint while globally adapting the other waypoints(Fig.\ref{fig:anecdotal_res} B).

\noindent\textbf{Evaluating Q2}: 
We use the Wilcoxon signed-rank test \cite{Wilcoxon} to assess response distribution and report the p-values. Participants noted that the framework better understood their preferences based on feedback (p<0.005), leading to higher success rates across all LLMs(Fig.\ref{fig:quant} D). Additionally, Participants reported a positive experience with the code-explanation feature(p<0.001), which provided clear insights into trajectory adaptations. This resulted in an impressive average interpretability score of 4.1 out of 5(Fig.\ref{fig:quant} E)

\noindent\textbf{Results against Baselines}
To benchmark against LaTTe and ExTraCT, we evaluated their methods with a similar user study to OVITA. For LaTTe, we employed their best pretrained model with n.depth=400 and set the locality factor to 0.5. For ExTraCT, we followed the paper’s settings: a deformation radius of 0.3 and weight w in the range [0.1,0.5]. Participants were not informed about the specific method used to deform the trajectories. Our method outperforms both LaTTe and ExTraCT on both the LaTTe subset (p<0.01) and extended subset (p<0.005 (Fig.\ref{fig:latte_compare}). A qualitative comparison of OVITA vs LaTTe vs ExTraCT can be seen in Fig.\ref{fig:latte_compare_qualitative}. Note that ExTraCT underperforms compared to LaTTe on the extended subset because it relies on discrete mappings, while LaTTe’s embedding-based architecture may capture parts of complex instructions. The mean inference time, measured over 30 samples, was approximately 12.3 seconds for OVITA, 3.2 seconds for LaTTe, and 0.4 seconds for ExTraCT.

\subsection{Simulation experiments}

\begin{figure}[t]
    \centering
    \includegraphics[width=0.95\linewidth]{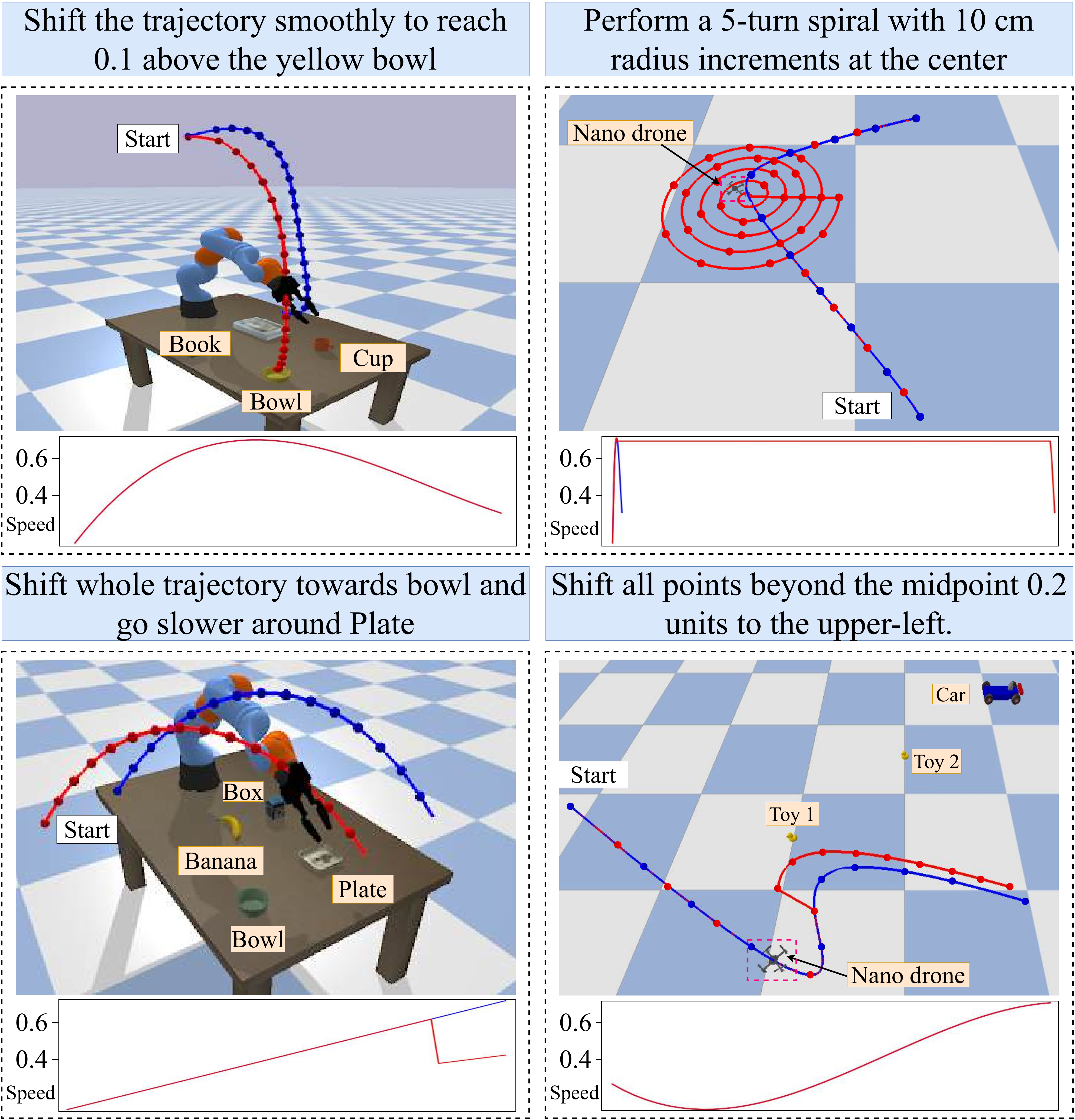}
    \caption{Simulation results with a robotic arm and drone along with speed profiles. Blue denotes the original trajectory, and red is the modified one.}
    \label{fig:sim_res}
    \vspace{-0.25in}
\end{figure}

\begin{figure} [t]
    \centering
    \includegraphics[width=0.95\linewidth]{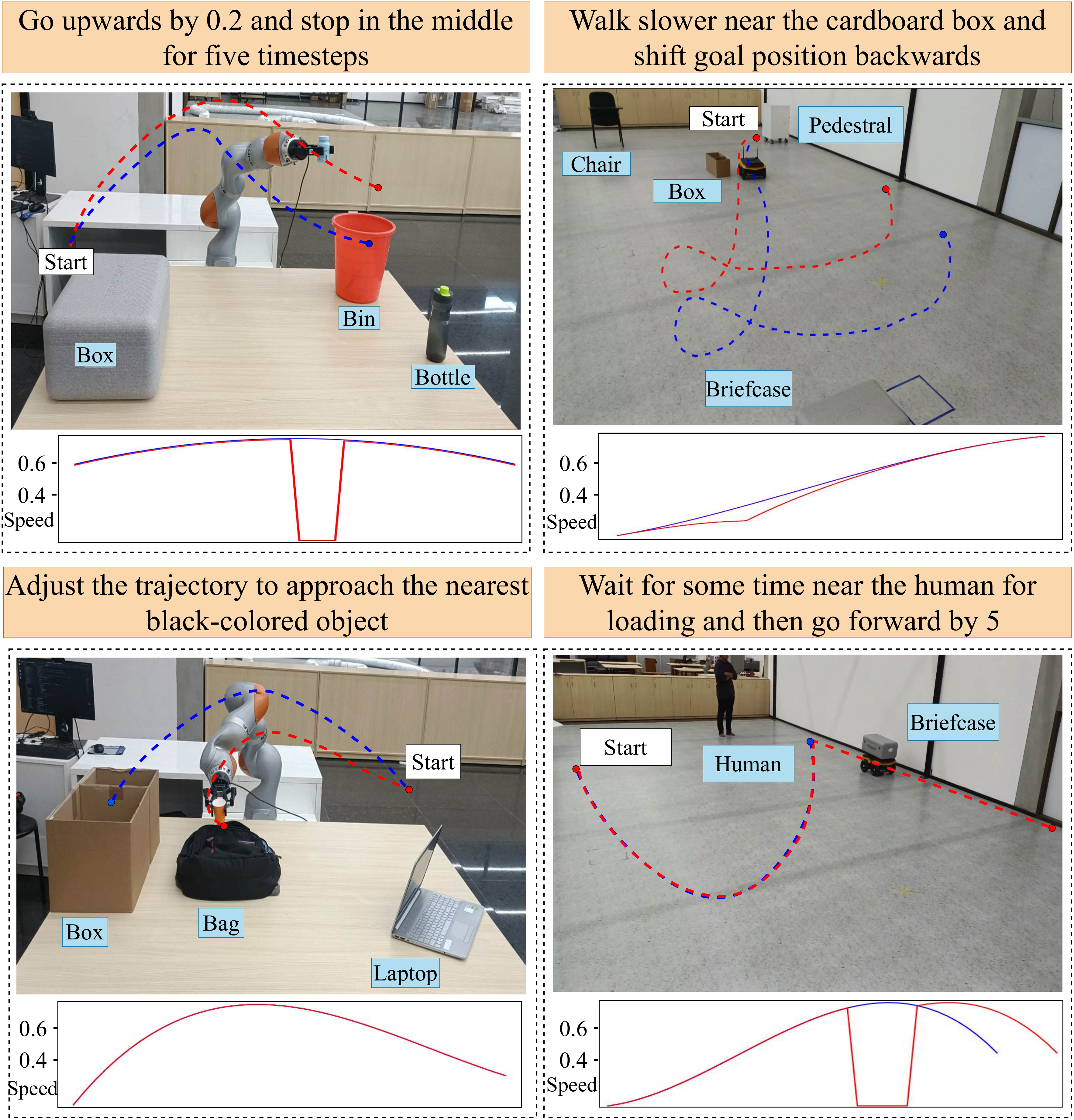}
    \caption{Real-World Testing on KUKA IIWA and Jackal Robots: Approximate trajectories are shown with original (blue) and modified (red) paths.}
    \label{fig:real_res}
    \vspace{-0.25in}
\end{figure}
To evaluate our approach across diverse robot dynamics and environments, we tested it on multiple robots and tasks. Experiments included a Crazyflie drone and a KUKA iiwa LBR 7-DOF robotic arm as shown in Fig.\ref{fig:sim_res}, demonstrating the framework's compatibility with various platforms, dynamics, and low-level IK modules. A 3D tabletop environment was created with varying object types and placements. The drone experiments focused on diverse motions in 3D navigation.

\subsection{Real-world experiments}
\label{real-world-experiments}
We validated our framework's executability on real-world robotic systems through experiments using the KUKA iiwa LBR 7-DOF manipulator and the Jackal ground robot, as shown in Fig.\ref{fig:real_res}. These trials focused on tasks exploring scenarios such as numerical translation of trajectory points, object-specific trajectory transformation, and multi-step deformation for robot operations. We used a standard CPU/GPU system connected via ROS to execute waypoints. {A mounted RGBD camera captures the workspace, and VLM provides an environment description, including object properties (color, shape, orientation, etc.). LangSAM (CLIP (ViT-L/14@336px)) \cite{C27} uses the image and object names to generate masks, which, combined with depth data and intrinsic camera parameters, yield 3D point clouds. From these, object centers, orientations (via PCA), and dimensions are computed in world coordinates.}

\subsection{Ablations}

We analyze our approach through ablation studies (Fig.~\ref{fig:quant}):  
\begin{itemize}
    \item \textbf{LLM Evaluation:} We compare multiple LLMs to identify the best-performing model and assess model dependence. Consistent results across multiple LLMs confirm our approach's generalizability (Fig.~\ref{fig:quant} A\&B).  
    \item \textbf{Impact of Human Feedback:} Incorporating feedback improves success rates across all LLMs, emphasizing its role in performance enhancement (Fig.~\ref{fig:quant} D).  
    \item \textbf{Code Executability:} Our framework generates 92.9\% executable code, demonstrating robustness in producing syntactically correct and actionable code. (Fig.~\ref{fig:quant} C).  
    \item \textbf{Interpretability:} High interpretability scores (avg. 4.1) show method transparency and compatibility with both expert and non-expert users. (Fig.~\ref{fig:quant} E).  

    \item {\textbf{Impact of example exclusion}: Success rates decline only slightly indicating strong robustness (Fig.~\ref{fig:quant} F).} 
    
        \item {We vary the LLM temperature (0-0.7) and report success rates in Fig.~\ref{fig:quant} G. Minimal variation shows robustness, with a sharp drop only at temperature 0.
}

\end{itemize}



\section{Conclusion and Limitations}
 OVITA successfully enables intuitive, language-based trajectory adaptation. A closed-loop feedback system ensures quality control and corrects LLM hallucination. The framework can be integrated with any existing waypoint-based planner, making it useful across diverse applications. An interesting application of our approach is natural language-driven trajectory augmentation, improving high-fidelity synthetic data generation. 
 
 Our approach, while effective in generating instruction-compliant trajectories, has several limitations. The LLM-generated code may fail due to syntax errors (e.g., incorrect function names), mathematical errors (e.g., division by zero), or deviations from the required function structure. Logical errors, though present, are readily corrected through feedback.  While OVITA aligns with user intent, it doesn’t guarantee globally optimal paths. Performance depends on precise user input, as vague terms (e.g., “very” or “relatively slower”) may be misinterpreted. Additionally, QP-based solvers can fail to find feasible solutions under overly restrictive or conflicting constraints. Future directions includes integrating generation with adaptation, extending to dynamic environments, and incorporating multimodal instructions (e.g., sketching via specialized simulators or AR/VR).
 








\bibliographystyle{IEEEtran}
\bibliography{IEEEexample}  

\end{document}